\documentclass[letterpaper]{article} 
\usepackage{aaai25}  
\usepackage{times}  
\usepackage{helvet}  
\usepackage{courier}  
\usepackage[hyphens]{url}  
\usepackage{graphicx} 
\usepackage{natbib}  
\usepackage{caption} 
 \usepackage{mathrsfs}
 \usepackage{amsmath}
\usepackage{algorithm}
\usepackage{algorithmic}
\usepackage{amssymb}
\usepackage{multirow}
\usepackage{newfloat}
\usepackage{listings}
\DeclareCaptionStyle{ruled}{labelfont=normalfont,labelsep=colon,strut=off} 
\floatstyle{ruled}
\newfloat{listing}{tb}{lst}{}
\floatname{listing}{Listing}
\title{Focus on Neighbors and Know the Whole: Towards Consistent \\ Dense Multiview Text-to-Image Generator for 3D Creation}
\author{
    Bonan Li\equalcontrib\textsuperscript{\rm 1,\rm 2},
    Zicheng Zhang\equalcontrib\textsuperscript{\rm 1},
    Xingyi Yang\textsuperscript{\rm 2},
    Xinchao Wang\textsuperscript{\rm 2}\thanks{Corresponding author}
}
\affiliations{
    \textsuperscript{\rm 1}University of Chinese Academy of Sciences, UCAS \ 
    \textsuperscript{\rm 2}National University of Singapore, NUS
    
}

\usepackage{bibentry}

\begin{document}

\maketitle

\begin{abstract}
Generating dense multiview images from text prompts is crucial for creating high-fidelity 3D assets. Nevertheless, existing methods struggle with space-view correspondences, resulting in sparse and low-quality outputs. In this paper, we introduce \textit{\textbf{CoSER}}, a novel \textbf{co}nsistent den\textbf{s}e Multivi\textbf{e}w Text-to-Image Gene\textbf{r}ator for Text-to-3D, achieving both efficiency and quality by meticulously learning neighbor-view coherence and further alleviating ambiguity through the swift traversal of all views. For achieving neighbor-view consistency, each viewpoint densely interacts with adjacent viewpoints to perceive the global spatial structure, and aggregates information along motion paths explicitly defined by physical principles to refine details. To further enhance cross-view consistency and alleviate content drift, CoSER rapidly scan all views in spiral bidirectional manner to aware holistic information and then scores each point based on semantic material. Subsequently, we conduct weighted down-sampling along the spatial dimension based on scores, thereby facilitating prominent information fusion across all views with lightweight computation. Technically, the core module is built by integrating the attention mechanism with a selective state space model, exploiting the robust learning capabilities of the former and the low overhead of the latter. Extensive evaluation shows that CoSER is capable of producing dense, high-fidelity, content-consistent multiview images that can be flexibly integrated into various 3D generation models.

\end{abstract}

\section{Introduction}

Creating 3D content from text is an important task in computer vision and graphics for its potential applications in advertising, robotics, digital games and the meta-verse.
Previous works~\cite{magic3d, fantasia3d,latent3d} primarily optimise 3D representations using score distillation sampling~\cite{dreamfusion} of Text-to-Image (T2I) models, ensuring that rendered images from any viewpoint maintain high fidelity while aligning with the given prompt.
Nevertheless, these methods suffer from Janus problems and severe artifacts with asymmetric geometry. As a remedy, multiview text-to-image (MT2I) models \cite{instant3d,v3d,vivid,epidiff,im3d,era3d} have been developed, providing rich 3D priors that help mitigate these problems. Furthermore, using dense views can accelerate the modeling process and enhance the quality of the resulting 3D models.

To acquire MT2I models, a promising approach is to leverage 3D priors learned from 3D datasets to enhance 2D diffusion models. Seminal works \cite{mvdream,wonder3d} have focused on inflating T2I models with dense cross-view attention to capture 3D information. However, this approach incurs expensive computation and faces challenges when dealing with dense viewpoints. To address these limitations, sparse attention mechanisms \cite{era3d} have been proposed to model cross-view dependencies, though they can introduce inconsistencies across multiple views. Recent studies \cite{videomv,sv3d,cat3d} have leveraged cross-frame continuity priors from text-to-video (T2V) models to enhance consistency. Nevertheless, azimuthal rotation around an object can cause significant content variations at the same image coordinates, making temporal attention insufficient for capturing intricate details necessary for 3D generation.

Despite impressive progress in multiview T2I generation, effectively and efficiently modeling the space-view structure within a unified architecture remains an open challenge. In this paper, we advance the field by addressing two fundamental problems:
(1) \textbf{\textit{How to design efficient cross-view interactions to ensure dense multiview consistency?}} Rather than directly optimizing consistency among all views with uniform granularity, we propose a two-step approach: (\emph{{\romannumeral1}}) Learning neighbor-view consistency as a prerequisite, as enforcing consistency between adjacent viewpoints naturally enables coherence across all viewpoints. 
(\emph{{\romannumeral2}}) Ensuring global-view coherence to mitigate discrepancies between distant views and facilitate unified appearance, since subtle inconsistencies are inevitable and deteriorated as the viewing angles increases.
\textbf{(2) \textit{Which operator should be applied for the implementation?}} 
Unlike previous works that employ attention as the default choice, we leverage hybrid operators, including attention~\cite{attention} and state space models (SSMs)~\cite{ssm}, to harness the benefits of both model families, \textit{e.g.}, the powerful capabilities of attention and the low computational cost of SSMs.

In this paper, we address these questions and propose a novel model named CoSER, which enhances consistency in dense MT2I tasks. CoSER utilizes ``factor'' interactions that focus on neighboring views and maintain awareness of the global view.
Technically, CoSER initially employs dense attention on adjacent views to capture the rough structure and texture of the current plane, thereby constructing a foundational appearance.
Subsequently, we introduce a trajectory-based module to refine local details by accurately tracing 3D trajectories and enhancing point-to-point consistency. Specifically, motivated by the efficiency of calculating rotated locations for centralized objects, we develop a robust method to handle complicate cases. This involves pixel interaction within a window of neighboring frames defined by physical rotation rules, ensuring precise alignment and consistency across views.
To further alleviate disparities across arbitrary viewpoints, we utilize state space models (SSMs) equipped with a novel spiral bidirectional scan strategy to aggregate features from the entire set of views. We then score each patch based on textual inputs and perform down-sampling conditioned on these scores to adaptively ignore semantically irrelevant information. This approach allows CoSER to efficiently maintain global context through sparse attention.
With these designs, our method establishes robust priors, enabling the creation of exceptionally coherent 3D assets (see Figure~\ref{fig:teaser}).

In summary, our contributions can be summarized as follows: 
\begin{itemize}
    \item We propose a novel multiview text-to-image framework, CoSER, for 3D content creation which exhibits significantly consistency behavior.
    \item We explore an efficient pipeline to model complex long-span consistency by focusing on neighbors to achieve short-span strict consistency and knowing whole views to obtain cues that correct cumulative errors.
    \item We design an hybrid architecture to implement the interaction by adopting both attention and SSMs as operators, effectively capturing essential dependencies in short sequences while significantly reducing costs for modeling long sequences.
    \item Extensive experiments demonstrate that our CoSER outperforms the state-of-the-art multiview synthesis approaches in both quantitative and qualitative results.
\end{itemize}

\section{Related Work}
\subsection{Text-to-3D Creation}
Recently, Text-to-3D generative models~\cite{tango, zerot, clipmesh,dreamfusion, bd23, 3dtopia} have made substantial advancements, achieving high-quality generation of both 3D objects and complex 3D scenes. Early works focus on directly generating assets using 3D diffusion models based on differentiable representations~\cite{pointe, rodin, sdfusion}. However, these approaches often relies on massive 3D data and incurs substantial training costs. Thanks to powerful text-to-image models~\cite{ldm}, DreamFusion~\cite{dreamfusion} leverages the robust 2D prior by introducing Score Distillation Sampling, which minimizes the discrepancy between rendered images from the underlying 3D assets and the diffusion prior. This development also sparked a surge of interest in other  generative tasks such as texture generation~\cite{textmesh, paint3d}, editing~\cite{ed, dreameditor, gaussianeditor, animatabledreamer, dreamcraft3d}, and Image-to-3D~\cite{make, imagedream, magic123}. Despite this paradigm has bridged the gap between imagination and reality, it still faces significant challenges in terms of optimization speed~\cite{hash3d,dreamtime}, content drift~\cite{prolificdreamer,lisweetdreamer} and the Janus problem~\cite{geodream,gaussiandreamer,seolet}. 
In response to inefficient training, ~\citet{latentnerf} train a NeRF within the latent space of LDM~\cite{ldm} to accelerate the learning process and further introduce the Sketch-Shape method to guide the generation process.
To acquire photorealistic appearance, ~\citet{magic3d} employ a coarse-to-fine strategy, first optimizing a NeRF using a low-resolution diffusion prior, then refining the texture with latent diffusion priors using the DMTet initialized with the coarse NeRF. ~\citet{fantasia3d} disentangle the learning of geometry and material, leveraging physics-based rendering techniques to achieve high-fidelity mesh generation. 
Another line of work lies in generating 3D assets directly through a 3D diffusion model based on differentiable representations. ~\cite{wonder3d} fine-tune image diffusion models to generate multiview image-normal pairs with switch attention and employ NeuS~\cite{neus} for reconstructin underlying geometries. In this work, we will continue to improve the quality of 3D content by designing the effective multiview text-to-image generator, which provides consistent images for reconstruction method.
\subsection{Multiview Image Generation}
Due to the scarcity of 3D data and the formidable training costs, lifting 2D diffusion to multiview generator conditioned on camera pose has become the prevailing paradigm~\cite{mvdream,wonder3d,instant3d,v3d,mvdiffusion,epidiff,im3d,era3d}. ~\citet{mvdiffusion} introduce the novel concept of generating multiview images concurrently with correspondence-aware attention mechanisms, thereby enabling effective cross-view information interaction and application to texturing scene meshes. Meanwhile, ~\citet{mvdream} and ~\citet{wonder3d} simply extends 2D spatial attention to capture 3D structure, yielding remarkable results. Nevertheless, the quadratic complexity along with resolution restricts the number of cameras, thereby compromising the performance of reconstruction algorithms and resulting in inferior results. To reduce computational costs, ~\citet{era3d} propose an efficient row-wise attention instead of dense interactions, yet struggles to correct local details referencing global information. Latter works~\cite{videomv,sv3d,cat3d} capitalized on priors derived from video diffusion model to achieve both dense view generation and consistent outputs. Despite these strides, a fundamental disparity persists between the video continuity and 3D consistency. General videos typically depict objects moving along a fixed plane from the same camera viewpoint, \textit{i.e.}, similar information shifts across different positions on the canvas. In contrast, videos composed of images captured from multiple views often exhibit significant rotation, leading to substantial observable content jitter between frames. Here, we tailor a innovative strategy for cross-view interaction that optimizes both performance and efficiency, ensuring robust consistency in generating dense-view images for 3D applications.
\begin{figure*}
\begin{center}
\includegraphics[width=1\linewidth]{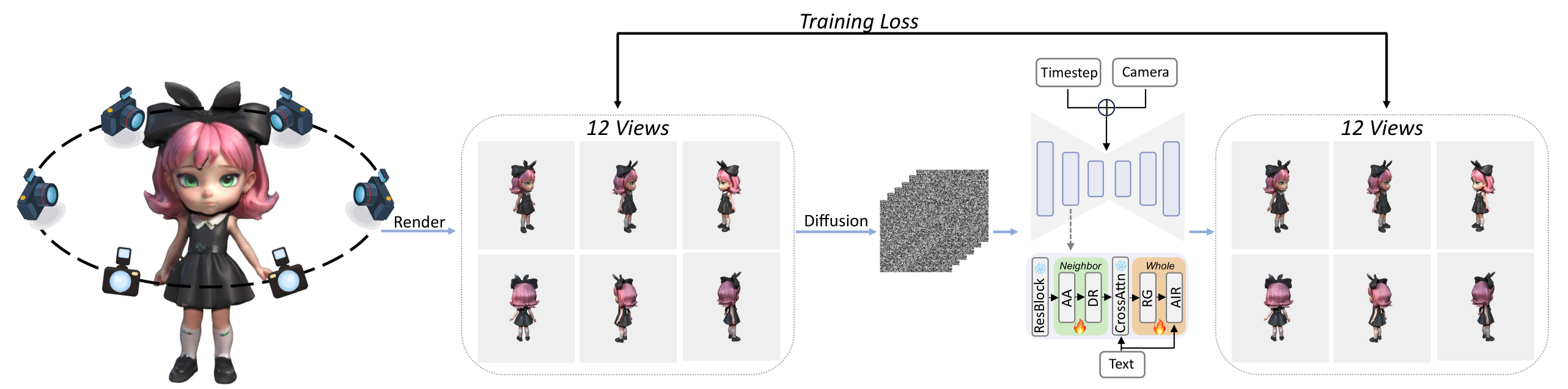}
\end{center}
    \vspace{-5mm}
   \caption{Illustration of our CoSER. Given images rendered from 12 views at the same elevation, we take a pre-trained text-to-image generation model and fine-tune it by incorporating camera poses and lifting 2D-UNet to generate multi-view images. Specifically, we achieve 3D perception by employing fine-grained learning for neighbors viewpoints and coarse-grained interactions across whole viewpoints. For neighbors, Appearance Awareness (AA) is used to learn the basic appearance and Detail Refinement (DR) is proposed to ensure neighbor consistency in point-level. For whole views, we quickly scan all viewpoints with Rapid Glance (RG) and then eliminate ambiguity with powerful Accumulated Inconsistency Rectification (AIR). }
   \vspace{-5mm}
\label{fig:pipline}

\end{figure*}
\section{Methodology}

\subsection{Preliminaries: Latent Diffusion Models}

Our work lifts the Latent Diffusion Model (LDM)~\cite{ldm}, which maps a noise latent $\epsilon$ and a text prompt $p$ to an image $x$. LDM first encodes the image $x_{0}$ into latent space using a pre-trained encoder $\mathcal{E}$, \textit{i.e.}, $z_{0} = \mathcal{E}(x_{0})$. The latent representation $z_{0}$ are then decoded back into the image $x_{0}$ using a pre-trained decoder $\mathcal{D}$.

The model is trained in the latent space in a self-supervised manner, and generates samples by reversing a Markov forward process. Specifically, the forward process injects noise into $z_{0}$ to obtain noisy inputs $\{z_{t}\}^{T}_{t=1}$, where $z_{t} = \sqrt{\alpha^{t}}z_{0} + \sqrt{1 - \alpha_{t}}\epsilon$, and $\epsilon \sim N(0,I)$. 

The generation process is formulated as a sequence of denoising operations $\epsilon_{\theta}(z_{t}, t, y)$, predicting a denoised variant of input $z_{t}$, where $y = \phi(p)$ is the condition, and $\phi$ is a frozen text encoder~\cite{clip}. 
The learnable parameter $\theta$ is typically embedded within structures such as convolutional layers, self-attention mechanisms, and cross-attention mechanisms. The training objective is defined as follows:
\begin{equation}
\label{eq:train}
\mathop{\mathrm{min}}_{\theta}{\mathbb{E}}_{z_{0}, \epsilon,t}||\epsilon-\epsilon_{\theta}(z_{t},t, y)||^2_2.
\end{equation}
The reverse process ensures that the final sample $z_{0}$ follows the distribution of latents, while aligning the generated image with the given condition.

\subsection{Overview of the CoSER Method}

We propose CoSER, a method for generating dense and consistent multiview images based on text prompts. The inputs to CoSER include a set of noisy images $\{x^{i}_{t}\}^{f}_{i=1}$, a set of camera parameters $\{c^{i}\}^{f}_{i=1}$ with a fixed elevation angle, uniformly distributed azimuth angles ranging from 0$^{\circ}$ to 360$^{\circ}$, and a text prompt $y$. We fine-tune the lifted Latent Diffusion Model (LDM) on $f$ views using a 3D dataset to synthesize multiview images $\{x^{i}_{0}\}^{f}_{i=1}$. 
It is non-trivial to incorporate the LDM architecture originally designed for 2D with 3D awareness. 
Rather than applying space-view attention or separately fusing spatial and viewpoint dimensions, our framework employs fine-grained interactions for neighboring views and coarse interactions for the global appearance. We retain the LDM structure, including ResBlock and CrossAttention, to maximize parameter reuse and maintain robust generative capabilities. A detailed sketch of CoSER is illustrated in Figure~\ref{fig:pipline} and please see more details for each module \textbf{in Appendix}.

\subsection{Focus on Neighbors with Contextual Attention}

It is well known that 3D objects exhibit strict cross-view consistency, whereas this does not necessitate considering the relationships among all views. For instance, the front and back views of an object are usually highly independent. Since adjacent viewpoints share more information and contribute significantly to visual consistency, we propose interacting each view with its context to mitigate the computational expense of space-view attention. Specifically, we use adjacent attention to capture the fundamental appearance and trajectory attention to refine details point-to-point, ensuring consistency between neighboring viewpoints. Contextual attention, in this case, refers to the mechanism of incorporating information from neighboring views to enhance the understanding of the current view.

\noindent\textbf{\textit{Appearance Awareness}} CoSER begins by integrating content from adjacent viewpoints to outline the basic appearance from the current view, thereby ensuring fundamental neighbor-view consistency. Following the MT2I baseline~\cite{mvdream}, we encode images $\{x^{i}\}$ to latent feature $ \{z^{i}\}$ by  the
pre-trained encoder $\mathcal{E}$ ($t$ is omitted for clarity), followed by mapping camera parameters $\{c^{i}\}$ as high-dimensional conditioin. Moreover, the original self-attention is replaced with adjacent-attention, which take which takes the previous view $z^{i-1}$, current view $z^{i}$ and next view $z^{i+1}$ as inputs, updating the features of $z^{i}$. The formulation of the
adjacent-attention is as follows:
\begin{equation}
    \begin{aligned}
Q^{i} &= W_{Q}z^{i}, \\
K^{i} &= W_{K}[z^{i-1},z^{i},z^{i+1}], V^{i} = W_{V}[z^{i-1},z^{i},z^{i+1}],
\end{aligned}
\end{equation}
where $[\cdot]$ denotes concatenation operation. To inherit the generalizability of the 2D diffusion models, the parameters of self-attention is used to initialize projection matrices $W_{Q}$, $W_{K}$, $W_{V}$ which are shared across space and view. Note that adjacent-frame attention, while capable of forming the initial appearance of the object, inevitably disperses attention due to the involvement of numerous interacting patches, making it struggle to focus on details.

\noindent\textbf{\textit{Detail Refinement}}
To ensure cross-view consistency in rendered images, it is crucial that the same position of an object corresponds to different pixels in neighboring viewpoints. These co-located pixels across views exhibit stronger correlations than others, playing a key role in maintaining visual consistency in multiview synthesis. For an object centered at the origin, and with some slight abuse of symbols, rigid rotation about the $y$-axis follows a physical principle:
\begin{equation}\label{transform}
    x'=(x-\frac{W}{2})\,\cos{\Delta \alpha}+\frac{W}{2}-d\,\sin{\Delta \alpha},
\end{equation}
where $(x, y)$ and $(x', y')$ are the pixel coordinates before and after rotation by an angle $\Delta \alpha$ at a depth $d$, and $W$ is the image width. Since the depth $d$ is unknown in the generative task and $\sin{\Delta \alpha}$ becomes negligible when $\Delta \alpha$ is small, we simplify the transformation to:
\begin{equation}\label{transform1}
x' = (x - \frac{W}{2})\cos{\Delta \alpha} + \frac{W}{2}.
\end{equation}
To make this simplification robust in practical applications, we consider a $3 \times 3$ window $\{x' + i, y' + j\}_{i,j = -1, 0, 1}$ in neighboring frames as potential results. The feature at pixel $(x, y)$ is then refined through the attention mechanism that interacts with these neighboring pixels, enhancing consistency across views.

\subsection{Know the Whole with Spiral Mamba}
To efficiently enhance global consistency, we first quickly scan all viewpoints, then down-sample patches based on textual relevance to filter out irrelevant information. This allows us to utilize powerful attention mechanisms to correct any accumulated inconsistencies.

\noindent\textbf{\textit{Rapid Glance}}
Exchanging information across all viewpoints is crucial for achieving global view consistency. However, dense attention mechanisms~\cite{wonder3d,mvdream} can be computationally expensive and memory-intensive, while simple temporal attention~\cite{videomv} fails to effectively learn multi-view consistency. Inspired by~\citet{ssm}, we leverage state space models (SSMs) known as linear operators to process all patches of multiview images, offering a more efficient alternative to quadratic-time attention. Specifically, we utilize the advanced selective scan SSMs introduced by Mamba~\cite{mamba} to implement our core SSMs operator. Details of Mamba can be seen \textbf{in Appendix}. Unlike existing Vision Mamba methods that flatten 2D images into 1D sequences row-wise or column-wise, we propose an effective spiral bidirectional scan strategy. This strategy leverages the prior knowledge that objects in 3D datasets are typically centered. By organizing object-related content within a short sequence window, our approach implicitly mitigates the forgetting problem of SSMs. Additionally, placing background content across distinct viewpoints aids the SSMs in boundary perception.

Technically, our method organizes spatial tokens starting from the center and expanding in a spiral pattern, then stacks them view by view. Meanwhile, we reverse the scan order within the arrangement of viewpoints to capture robust space-view features. This allows us to perform a rapid glance as follows:
\begin{equation}
    \{z^{i}\}^{f}_{i=1} = SSMs(SBScan( \{z^{i}\}^{f}_{i=1})), 
    \label{equ:adj_attn}
\end{equation}
where $SBScan(\cdot)$ denotes the spiral bidirectional scan. This operator effectively aggregates global information and provides crucial clues for further disambiguation.

\noindent\textbf{\textit{Accumulated Inconsistency Rectification}} To facilitate the elimination of inconsistencies through attention mechanisms, we implement down-sampling to filter out background and retain features relevant to textual information, thereby substantially reducing the number of patches engaged in interactions. Specifically, we provide a high-level semantic embedding $\phi(p)$ of the prompt $p$. A 2-layer MLP $S$ is employed to assess the relevance between each patch and the text, producing a score ranging from 0 to 1 at each position, where a higher score indicates stronger relevance. Based on the score map $s^{i} = S([z^{i},\phi(p)])$ as a weight, we perform 2D-pooling along the spatial dimensions on the query, key, and value maps to enable sparse interactions:
\begin{equation}
    \begin{aligned}
\hat{Q}^{i} &= P(s^{i}\cdot \bar{Q}^{i};\tau),\\ 
\hat{K}^{i} &= P(\{s^{j}\cdot \bar{K}^{i}\}^{f}_{j=1};\rho),\\ 
\hat{V}^{i} &= P(\{s^{j}\cdot \bar{V}^{i}\}^{f}_{j=1};\rho).
\end{aligned}
\end{equation}

Here, $P$ denotes the pooling operator, $\tau$ and $\rho$ are the strides required for pooling, and the transformations include: $\bar{Q}^{i} = W_{\bar{Q}}z^{i}$, $\bar{K}^{i} = W_{\bar{K}}[\{z^{j}\}^{f}_{j=1}]$, and $\bar{V}^{j} = W_{\bar{V}}[\{z^{j}\}^{f}_{j=1}]$. Notably, in contrast to the intensive down-sampling of $\hat{V}^{i}$ and $\hat{K}^{i}$, we maintain a higher resolution for the query feature map $\hat{Q}^{i}$ to minimize the loss of content information. Finally, we up-sample the computed feature map using bilinear interpolation to restore its resolution.

\begin{figure*}
\begin{center}
\includegraphics[width=1\linewidth]{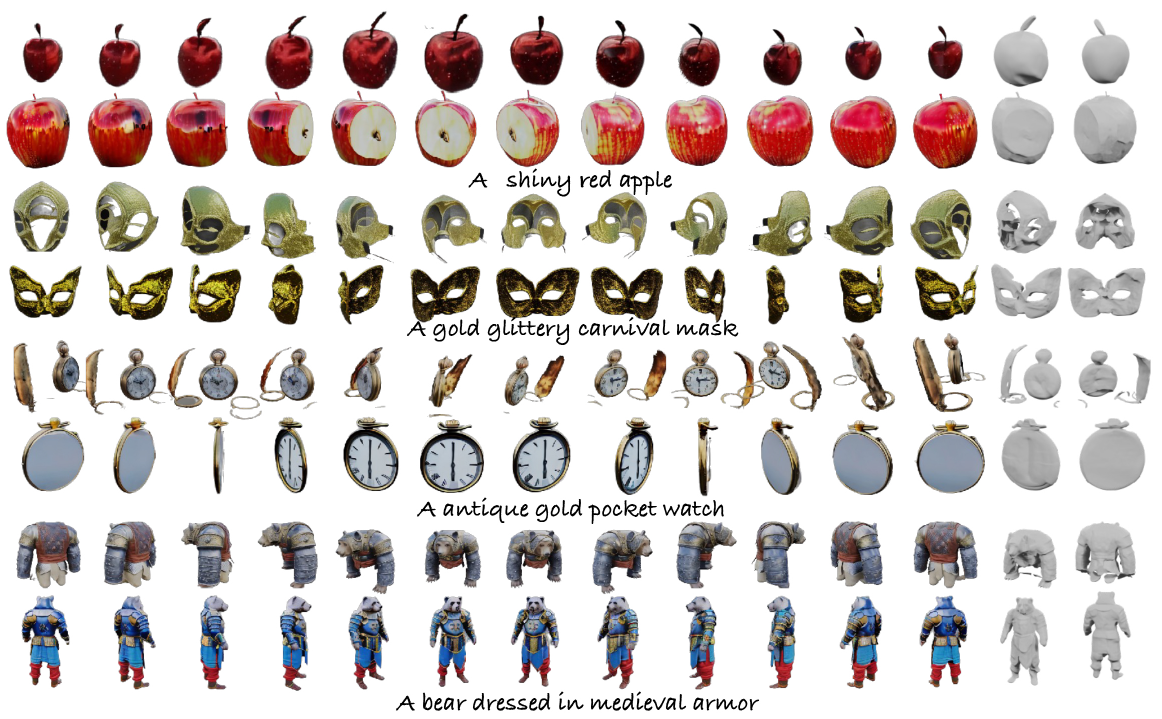}
\end{center}
 \vspace{-2mm}
   \caption{Qualitative comparison of VideoMV~\cite{videomv} (Up) and our CoSER (Down).}
\label{fig:main_1}
\end{figure*}

\begin{figure*}
\begin{center}
\includegraphics[width=1\linewidth]{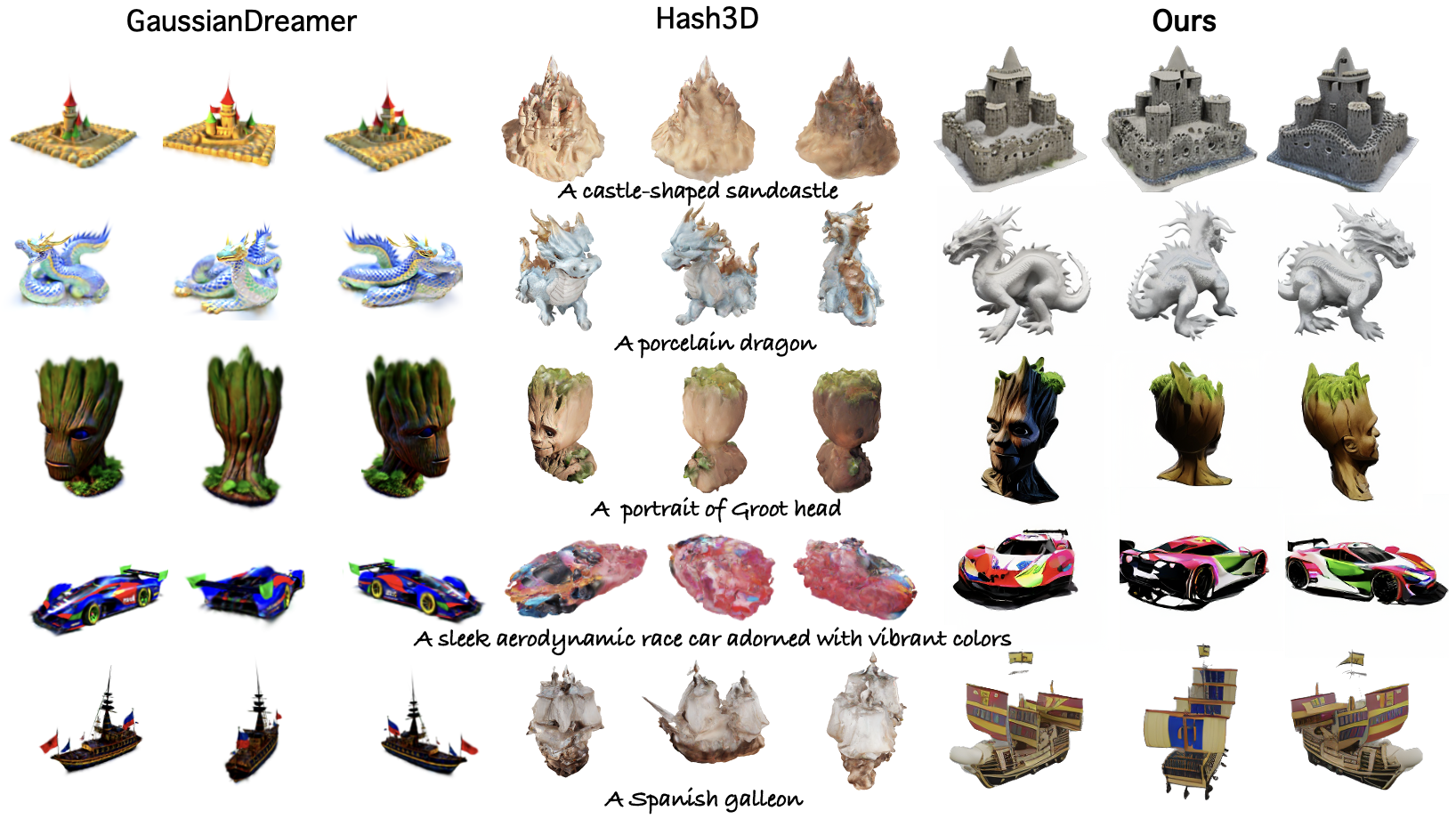}
\end{center}
\vspace{-2mm}
   \caption{Qualitative comparisons of GaussianDreamer~\cite{gaussiandreamer}, Hash3D~\cite{hash3d} and our CoSER.}
\label{fig:main_2}
\end{figure*}
\vspace{-5mm}
\section{Experiments}
In this section, we first provide the implementation details and the dataset. We then conduct both qualitative and quantitative experiments to compare our method with previous state-of-the-art text-to-3D generation methods. Additionally, we perform an ablation study to demonstrate the effectiveness of the proposed modules. 

\noindent\textbf{\textit{Implementation}} 
We adopt the LDM\footnote{\url{https://github.com/CompVis/stable-diffusion}}~\cite{ldm}, pre-trained on the LAION-5B~\cite{LAION-5B}, as our base model. Specifically, we use the AdamW optimizer~\cite{adamw} with a learning rate of 3e-5 and a batch size of 6 to optimize the weights over approximately 200K iterations on three NVIDIA Tesla A800 GPUs. To preserve diversity and detail, we degrade CoSER to a text-to-image generation model with a 40\% probability and train it with a subset of LAION-5B. During inference, we employ the DDIM sampler with 50 steps and set the scale guidance to 7.5 for generation. Additionlly, we set text template as ``A DSLR photo of [prompt], 3d asset'' to acquire more realistic results. Following the \citet{syncdreamer}, we use NeuS~\cite{neus} for performing reconstruction.

\noindent\textbf{\textit{Dataset}} 
We use the public G-Objaverse~\cite{richdreamer} as the training dataset, utilizing the names of the objects appended with ``3D asset'' as their text labels. For each object, we first normalize the object to be centered within a bounding box ranging from [-0.5, 0.5]. The camera distance is set to 2.0 units. A random elevation is chosen between [-30$^\circ$, 30$^\circ$], and starting from the front view, we uniformly sample 12 angles around the object for rendering.

\begin{table}
\scriptsize
\begin{center}
\begin{tabular}{ccccc}
\hline
\multirow{2}{*}{Model} & \multirow{2}{*}{CLIP-S$\uparrow$} & \multicolumn{3}{c}{User Study} \\

& & Semantic$\uparrow$  & Consistency$\uparrow$ & Texture$\uparrow$  \\ \hline

GaussianDreamer &  30.78  & 0.26   & 0.27     & 0.21    \\
Hash3D          &31.26     & 0.20   & 0.19     & 0.25    \\
VideoMV         & 32.19    & 0.23    & 0.25    & 0.22     \\
\textbf{Ours} & \textbf{33.07}   & \textbf{0.31}   & \textbf{0.29}   & \textbf{0.32}   \\
\hline
\end{tabular}
\caption{Quantitative comparison of CoSER and state-of-the-arts based on the CLIP Score and User Study.}
\label{tab:user}
\end{center}
\vspace{-10mm}
\end{table}

\subsection{Comparison with SOTA methods}
\noindent\textbf{{\textit{Qualitative Comparison}}} 

\begin{figure*}
\begin{center}
\includegraphics[width=1\linewidth]{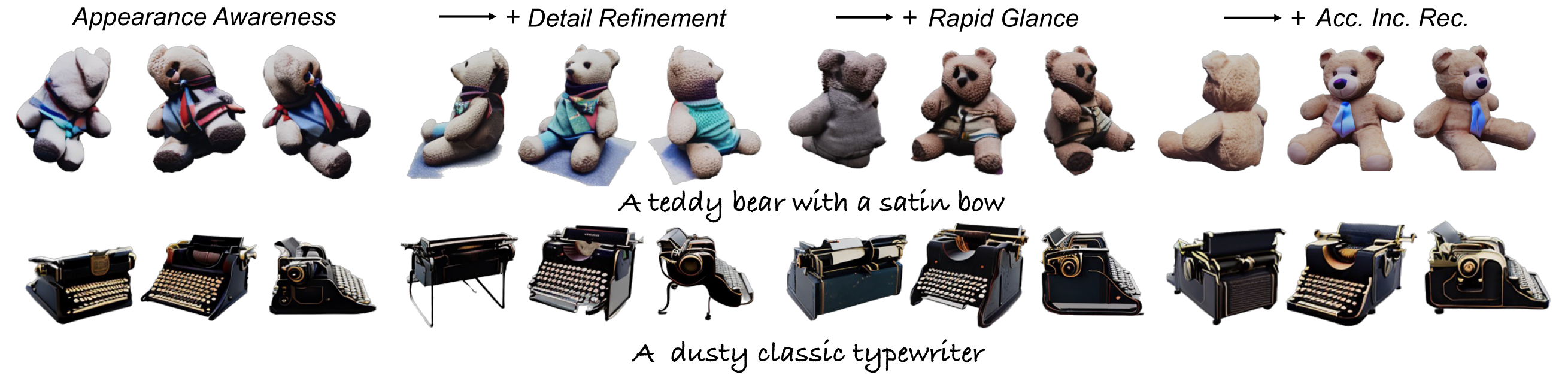}
\end{center}
\vspace{-5mm}
   \caption{Ablation of proposed moudles. The arrows indicate the changes made to the Appearance Awareness by sequentially plusing Detail Refinement, Rapid Glance, and Accumulated Inconsistency Rectification. Better see in color and 2x zoom.}
\label{fig:aba_1}
\vspace{-5mm}
\end{figure*}

We primarily evaluate the performance of the proposed CoSER on multiview text-to-image and compare the results with state-of-the-art methods in Figure~\ref{fig:aba_1}. For VideoMV, 12 images generated from the same camera viewpoints as CoSER are selected for visualization, and all 24 viewpoints were utilized in the reconstruction process to ensure a fair comparison. Qualitatively, our method demonstrates impressive quality improvement in terms of both consistence and texture. Although VideoMV can generate images from a broader range of perspectives, \textit{e.g.,} mask and watch, it significantly compromises cross-view consistency, which poses serious challenges for subsequent 3D reconstruction. Furthermore, Figure~\ref{fig:main_2} presents a comparison between two state-of-the-art text-to-3D methods, including GaussianDreamer~\cite{gaussiandreamer} and Hash3D~\cite{hash3d}. Specifically, the advanced method, DreamGaussian~\cite{dreamgaussian}, is utilized as the foundational model for Hash3D. Thanks to the dense consistent images, CoSER outperforms other methods in terms of fidelity in 3D geometry and visual appearance.

\begin{figure}
\begin{center}
\includegraphics[width=1\linewidth]{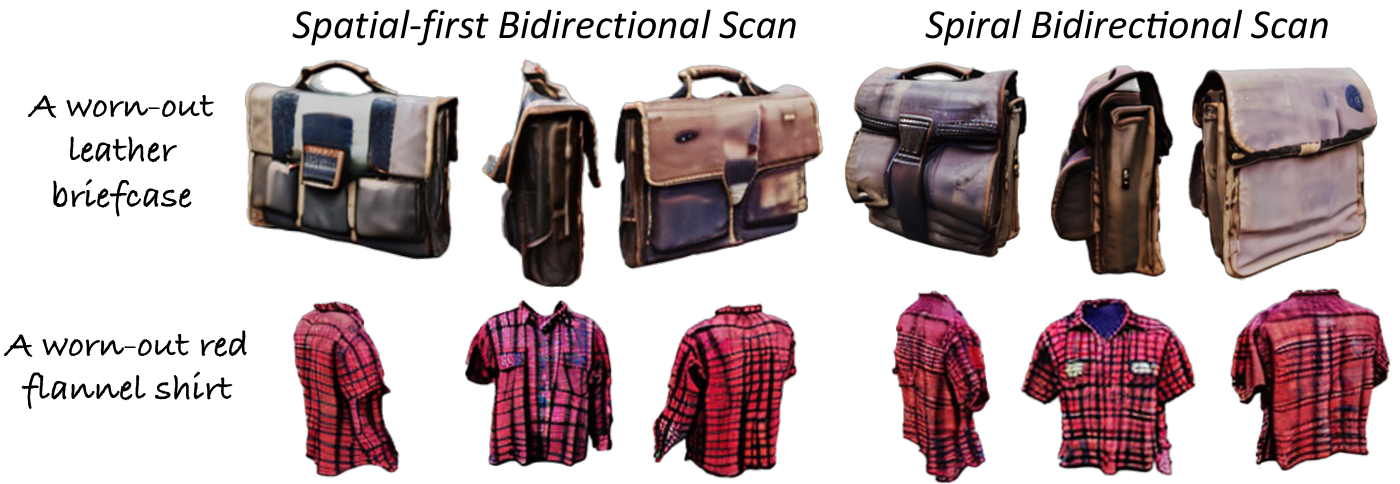}
\end{center}
\vspace{-5mm}
   \caption{Ablation of scanning strategy. Spiral bidirectional scan generates semantica- and view-consistent images.}
\label{fig:aba_2}
\vspace{-6mm}
\end{figure}

\noindent\textbf{{\textit{Quantitative Comparison}}} Following previous work~\cite{videomv}, we quantitatively assess the quality of the generated 3D assets using CLIP Score~\cite{clips}. We employ 100 text prompts selected from T3Bench~\cite{T3} to generate 3D models. For each object, we set the elevation as 15$^\circ$ and select 120 evenly spaced azimuth angles from -180$^\circ$ to 180$^\circ$, thereby generating 120 rendered images from different viewpoints. During the evaluation, 30 from the 120 rendered images are randomly selected and the CLIP Score is computed as the average semantic similarity between the prompts and these images. We use the CLIP ViT-B~\cite{clipvit} to extract both text and image features. The results, presented in Table~\ref{tab:user}, demonstrate that our method significantly surpasses other methods, indicating that CoSER is more adept at injecting semantic information into the creation of realistic 3D assets. Given the challenges in automating the measurement of 3D consistency, we resort to user studies to assess which method produces results that are most favored by human evaluators. We conducted three user studies on the results in terms of the Semantic Alignment, Multiview Consistency and Texture Detail. In the first study, participants were asked to select the 3D asset that best aligns with the semantics of the given prompt. In the second study, participants identified the asset with the highest consistency. In the third study, participants were encouraged to choose the object with the finest texture and less distortion artifacts. We receive 100 answers on 30 text-3D pairs in total for each study. As shown in Table~\ref{tab:user}, over 29\% of our results are selected as the best in both three metrics, which proves a significant advantage in 3D generation.

\subsection{Ablation Study}

\begin{figure}
\begin{center}
\includegraphics[width=1\linewidth]{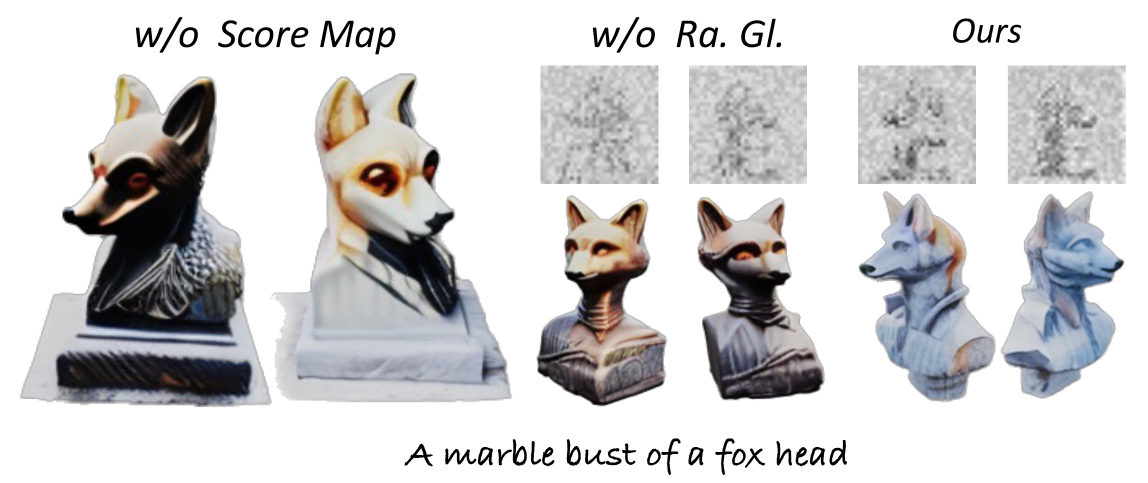}
\end{center}
\vspace{-5mm}
   \caption{Ablation of score map. Score map aids in capturing critical information, while Rapid Glance contributes to the generation of more accurate score maps.}
\label{fig:aba_3}
\vspace{-8mm}
\end{figure}

\noindent\textbf{{\textit{Effect of Module Design}}}  We are mainly interested in the 3D-aware module design, given that other unit has been well explored in previous methods. The core of our design is focusing on neighbors in fine-grained manner and knowing whole views with coarse-grained mechanism. To substantiate these claims, we incrementally integrated the proposed modules into four distinct configurations and fine-tuned the LDM accordingly. As depicted in Figure~\ref{fig:aba_1}, the use of Appearance Awareness alone proves insufficient for ensuring cross-view consistency. However, the incorporation of Detail Polishing enhances the alignment between adjacent views, thereby improving overall 3D content consistency. It is evident that Rapid Glance offers small reduction in ambiguities; nonetheless, its primary benefit lies in facilitating information exchange across all viewpoints (See Figure~\ref{fig:aba_3}), thereby establishing a foundation for the effective Accumulated Inconsistency Rectification module. 

\noindent\textbf{{\textit{Spiral Bidirectional Scan in Rapid Glance}}} To verify the effectiveness of the proposed spiral bidirectional scan strategy, we perform the evaluation by replacing it with vanilla spatial-first bidirectional scan in VideoMamba~\cite{videomamba}. As shown in Figure~\ref{fig:aba_2}, both scanning strategies enhance the consistency. However, spatial-first bidirectional scan inevitably disrupts the continuity of object, which hampers the efficiency of SSM model for feature aggregation. In contrast, our proposed spiral bidirectional scan maximizes the spatial compactness of foreground content while simultaneously enforcing operator awareness of inter-view boundaries, thereby achieving better results. Additionally, we experimented with reversing the spiral scanning order within the image, progressing from the periphery inward towards the center. However, this strategy did not yield further performance improvements, which we attribute to the already efficient information aggregation within an image achieved by the  spiral scan.

\noindent\textbf{{\textit{Score Map in Accumulated Inconsistency Rectification}}} In Accumulated Inconsistency Rectification, we introduce the score map to assist in downsampling feature resolution for coarse attention, tailored to each individual view. To understand its impact, we compared the full AIR module generated results with those from two alternative models: AIR without score map and AIR with score map but without Rapid Glance. For clearness, we extract score maps at various resolutions and upsample them to 32x32 for visualization. As shown in Figure~\ref{fig:aba_3}, the score map assigns higher scores to foreground regions, while background areas receive lower scores, facilitating the retention of critical information during the downsampling process. Additionally, Rapid Glance enhances the exchange of information, contributing to the generation of more precise score maps.

\vspace{-3mm}
\section{Conclusion}

In this paper, we introduced \textit{\textbf{CoSER}}, a novel framework for generating consistent and high-fidelity dense multiview images from text prompts. Our approach leverages a hybrid architecture that encapsulate the strengths of attention mechanisms and state space models (SSMs) to address the crucial challenges of cross-view consistency. 
The innovative design of contextual interaction includes an adjacent attention for capturing rough structures and textures, a trajectory-based attention for refining local details. A spiral bidirectional Mamba ensures global coherence in a cost-efficient manner. This comprehensive approach allows CoSER to produce dense, high-fidelity, and content-consistent multiview images.
Extensive experiments demonstrate that CoSER outperforms state-of-the-art multiview synthesis methods in both quantitative and qualitative evaluations across various datasets. 
Future work could further explore the hybrid way of neighbor and global interaction to efficiently generate dense and consistent multiview images.

\bibliography{aaai25}

\end{document}